%% file: 3677.tex
\documentclass[runningheads]{llncs}
\usepackage{graphicx}

\usepackage{tikz}
\usepackage{comment}
\usepackage{amsmath,amssymb} 
\usepackage{color}

\usepackage[accsupp]{axessibility}  

\usepackage{booktabs}
\usepackage{multirow}
\usepackage{mathrsfs}
\usepackage{mmstyles}
\usepackage{marvosym}
\usepackage{diagbox}
\newcommand{\rev}[1]{\textcolor{black}{#1}}

\begin{document}
\pagestyle{headings}
\mainmatter
\def\ECCVSubNumber{3677}  
\title{Counterfactual Intervention Feature Transfer for Visible-Infrared Person Re-identification} 
\titlerunning{Counterfactual Intervention Feature Transfer for VI-ReID}
%
\author{
Xulin Li\inst{1,2*} \and
Yan Lu\inst{1,2*} \and
Bin Liu\inst{1,2}\textsuperscript{\Letter} \and
Yating Liu\inst{3} \and
Guojun Yin\inst{1,2} \and
Qi Chu\inst{1,2} \and
Jinyang Huang\inst{1,2} \and
Feng Zhu\inst{4} \and
Rui Zhao\inst{4,5} \and
Nenghai Yu\inst{1,2}
}
\authorrunning{X. Li et al.}
%
\institute{
School of Information Science and Technology, University of Science and Technology of China \and
Key Laboratory of Electromagnetic Space Information, Chinese Academy of Science \and
School of Data Science, University of Science and Technology of China \and
SenseTime Research \and
Qing Yuan Research Institute, Shanghai Jiao Tong University\\
\email{\{lxlkw,luyan17\}@mail.ustc.edu.cn},
\email{flowice@ustc.edu.cn}\\
\email{\{liuyat,gjyin\}@mail.ustc.edu.cn},
\email{qchu@ustc.edu.cn}\\
\email{huangjy@mail.ustc.edu.cn},
\email{\{zhufeng,zhaorui\}@sensetime.com}\\
\email{ynh@ustc.edu.cn}
}
\maketitle
\renewcommand{\thefootnote}{*}
\footnotetext{Equal contribution.}
\renewcommand{\thefootnote}{\Letter}
\footnotetext{Corresponding authors.}

\input{./contents/abstract}
\input{./contents/introduction}
\input{./contents/relatedwork}
\input{./contents/review}
\input{./contents/method}
\input{./contents/experiments}
\input{./contents/conclusion}

\input{./contents/prove}
\input{./contents/loss}
\input{./contents/gft}
\input{./contents/vis}
\clearpage
%
%
\bibliographystyle{splncs04}
\bibliography{references}
\end{document}

%% file: contents/abstract.tex
\begin{abstract}

Graph-based models have achieved great success in person re-identification tasks recently,  which compute the graph topology structure (affinities) among different people first and then pass the information across them to achieve stronger features. But we find existing graph-based methods in the visible-infrared person re-identification task (VI-ReID) suffer from bad generalization because of two issues: 1) \textbf{train-test modality balance gap},
which is a property of VI-ReID task. The number of two modalities data are balanced in the training stage but extremely unbalanced in inference, causing the low generalization of graph-based VI-ReID methods. 
2) \textbf{sub-optimal topology structure} caused by the end-to-end learning manner to the graph module. We analyze that the joint learning of backbone features and graph features weaken the learning of graph topology, making it not generalized enough during the inference process.
In this paper, we propose a Counterfactual Intervention Feature Transfer (CIFT) method to tackle these problems. Specifically, a Homogeneous and Heterogeneous Feature Transfer (H$^2$FT) is designed to reduce the train-test modality balance gap by two independent types of well-designed graph modules and an unbalanced scenario simulation. Besides, a Counterfactual Relation Intervention (CRI) is proposed to utilize the counterfactual intervention and causal effect tools to highlight the role of topology structure in the whole training process, which makes the graph topology structure more reliable.
Extensive experiments on standard VI-ReID benchmarks demonstrate that CIFT outperforms the state-of-the-art methods under various settings.

\keywords{Person Re-identification, Counterfactual, Cross-modality}

\end{abstract}

%% file: contents/introduction.tex
\section{Introduction}
\label{sec:intro}

\begin{figure*}[t]
    \centering
    \includegraphics[width=0.95\linewidth]{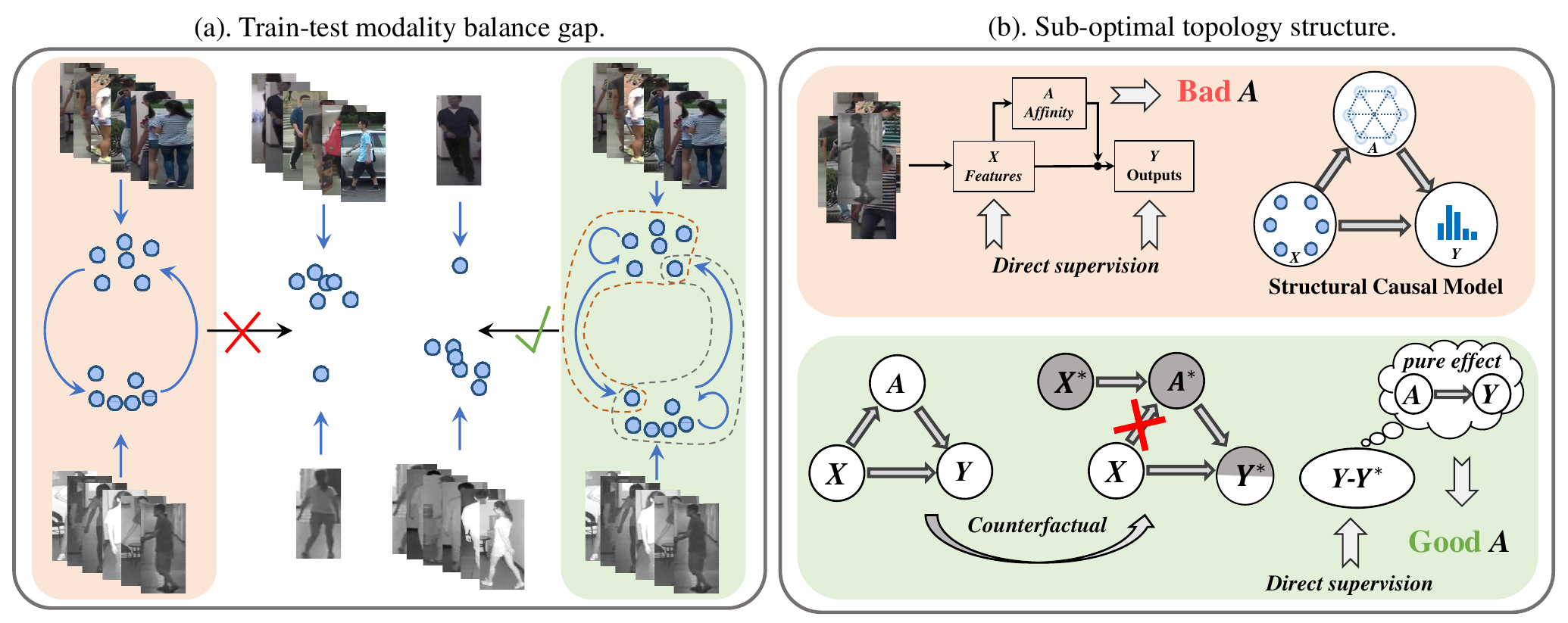}
    \caption{
    The red background hints at the existing methods and the green one means our method. (a) Existing graph-based methods trained on modality-balanced data are difficult to transfer to a modality unbalanced scenario. Our method overcomes this problem by unbalanced scenarios simulation and a novel graph module design. 
    (b) Existing training strategies learn the backbone features and the graph outputs jointly, making affinity learning be weakened. Our method uses a counterfactual intervention tool to calculate the pure effect contributed by the affinity changes only, making the model perceive the role of graph topology more direly.
    }
    \hfill
    \label{fig:1}
\end{figure*}

Standard person re-identification (ReID)~\cite{gong2014re,luo2019bag,sun2020circle,sun2018beyond,zhang2017alignedreid} aims to match pedestrian images of the same identity captured by different cameras, which is essentially a single-modality (RGB) retrieval task.
However, open-world intelligent monitoring requires methods to retrieve targets captured by infrared or thermal cameras in the dark scenario. Therefore, the research on visible-infrared person re-identification (VI-ReID)  has attracted great attention in recent years. Different from standard ReID, large cross-modality discrepancy and intra-modality variations bring new challenges to research on VI-ReID. 
Most researchers~\cite{hao2021cross,liu2020parameter,wu2021discover,ye2018hierarchical,ye2018visible,zhang2021global,zhu2020hetero} aimed to embed images of two modalities into the same feature space to tackle this task, which preliminarily solved the modality gap. 

Graph-based methods have achieved excellent performance on standard person ReID~\cite{chen2018group,luo2019spectral,shen2018deep,shen2018person}.
Generally, they predict the pair-wise similarity as relationships between different samples and then utilize those relationships to propagate messages across samples. This kind of methods can bring a large range of performance gain because the features of one sample not only have the discriminative information of this sample itself but also carry information from other relative samples. So several methods~\cite{lu2020cross,ye2020dynamic} attempt to employ graph-based modules to establish relationships and enhance features in the VI-ReID task. But we argue that different from the graph module on the standard Re-ID, the graph methods on VI-ReID suffer bad generalization. 

We delve into the graph models pipeline in VI-ReID and summarize two main reasons for the bad generalization: \textit{Train-test modality balance gap} and \textit{Sub-optimal topology structure}. 
The train-test modality balance gap is a property of VI-ReID task, which means that the number of two modalities data are balanced in the training stage but extremely unbalanced in inference, like Fig.~\ref{fig:1} (a) shows.
More details about this property will be further introduced in Section~\ref{sec:3}. 
The sub-optimal structure is another problem but usually ignored by previous methods. 
It is caused by the end-to-end learning manner of the graph model. We summarize that the existing joint learning process of both the backbone and the graph module would weaken the learning of the graph topology structure, like Fig.~\ref{fig:1} (b) shows.
They both lead to low generalization of the graph structure predicted by the graph module in inference. 

To tackle the aforementioned problems respectively, we propose a Counterfactual Intervention Feature Transfer (CIFT) including one new graph module called Homogeneous and Heterogeneous Feature Transfer (H$^2$FT) with one additional learning methods Counterfactual Relation Intervention (CRI). 
The H$^2$FT aims to reduce the train-test modality balance gap in two ways, training algorithm and model designing. We reorganize the balanced training data to simulate unbalanced modality distributed scenarios and let the H$^2$FT trained on that environment, which guides the model to adapt to the situation with unbalanced modality distribution. Also, we find that it is hard for the standard graph module to train efficiently on that unbalanced data because the standard message-passing process cannot adapt to the extremely unbalanced modality information. So, we carefully construct the module of the H$^2$FT which includes two different types of graph modules, to reduce the useless information introduced by the standard graph module and treat the message passing in unbalanced data better, as Fig.~\ref{fig:1} (a) shows. 
Except that, the CRI tackles the sub-optimal graph topology problem by highlighting the role of graph structure (predicted affinity) in the total end-to-end training. 
We utilize the tools of causal inference to implement that motivation. We first represent our graph module in the Structural Causal Model~\cite{pearl2016causal,pearl2018book} in Fig.~\ref{fig:1} (b) and modify the training targets of the graph module from only maximizing the probability likelihood to maximizing the combination of both probability likelihood and the total indirect effect (TIE). The former term guides the whole model to classify the identity of each person image. And the latter one is essentially equal to maximize the difference between the original output and a counterfactual output contributed by the affinity changes only (Fig.~\ref{fig:1} (b) green background), which can make the model perceive the function of the graph affinity. 

The main contributions of our work are summarized as follows:

\noindent $\bullet$ We delve into the existing VI-ReID graph model and find two main reasons for their low generalization: train-test modality balance gap and sub-optimal structure. And we design a novel and effective Counterfactual Intervention Feature Transfer (CIFT) to tackle these problems and achieve the new state of the art.

\noindent $\bullet$ We introduce a Homogeneous and Heterogeneous Feature Transfer (H$^2$FT) module including two independent types of well-designed graph module and an unbalanced scenario simulation, which is more suitable for tackling the sample interaction in the scenario with unbalanced modality distribution. 

\noindent $\bullet$ We propose a novel Counterfactual Relation Intervention (CRI) algorithm to tackle the sub-optimal topology structure problem. It utilizes the counterfactual intervention and causal effect tools to highlight the role of the topology in the feature transfer module, which can train the total module more generalized.

%% file: contents/relatedwork.tex
\section{Related Work}
\label{sec:relate}

\noindent \textbf{Visible-Infrared Person Re-ID.} Traditional single-modality person Re-ID ~\cite{luo2019bag,sun2018beyond,zhang2017alignedreid} is limited by the poor illumination conditions at night, so the VI-ReID has received extensive attention in recent years. Many VI-ReID approaches have been proposed to overcome the modality discrepancy produced by different cameras. Wu \etal.~\cite{wu2017rgb} proposed a deep zero-padding network and contribute the first large-scale multiple modality Re-ID dataset named SYSU-MM01. 

Many works~\cite{hao2021cross,ling2021multi,liu2020parameter,wu2021discover,ye2018hierarchical,ye2018visible,zhang2021global,zhu2020hetero} designed loss functions from the perspective of metric learning to better embed different modalities into the same feature space. Zhu \etal.~\cite{zhu2020hetero} proposed the hetero-center loss to reduce the intra-class cross-modality variations. Liu \etal.~\cite{liu2020parameter} proposed the hetero-center triplet loss to relax the strict constraint of traditional triplet loss. 

Some methods~\cite{dai2018cross,wang2020cross,wang2019rgb,wang2019learning} are based on the generative adversarial network (GAN)~\cite{goodfellow2020generative}. cmGAN~\cite{dai2018cross} adopted generative adversarial training to better distinguish images of different modalities at the feature level. D$^2$RL~\cite{wang2019learning} applied dual-level discrepancy reduction learning based on a bi-directional cycle GAN. Recently, Wu \etal.~\cite{wu2021discover} introduced a modality alleviation module and a pattern alignment module to discover cross-modality nuances. Hao \etal.~\cite{hao2021cross} confused two modalities, ensuring that the optimization is explicitly concentrated on the modality-irrelevant perspective. 
All these methods treat the VI-ReID as an image embedding task and learn to 
extract features directly from the single image.

\noindent \textbf{Graph-based Person Re-ID.}
In the single-modality person Re-ID task, except for the image embedding method, some works~\cite{chen2018group,luo2019spectral,shen2018person,shen2018deep} pay attention to the relationship between sample pairs. These methods introduced more supervised information of graph relationships into the training stage, while the inference stage also benefits from pair-wise similarity.
Besides, some re-ranking~\cite{zhong2017re} and graph neural networks (GNN)~\cite{zhang2020understanding} methods only treated relational modeling as post-processing to more flexibly adapt to various backbone networks. 

In VI-ReID, the large cross-modality discrepancy makes the optimization of the relationship more difficult. Recently, some approaches have explored cross-modality pair-wise relation learning with graph networks. Ye \etal.~\cite{ye2020dynamic} introduced cross-modality graph-structured attention to enhance robustness against noisy samples. Lu \etal.~\cite{lu2020cross} proposed the cross-modality shared-specific feature transfer algorithm that utilizes the graph convolution operator to propagate features over a graph to supplement the information of another modality. These methods utilized the graph network or transformer module to propagate message cross samples to extract stronger features. But they are all suffering from the train-test modality balance gap and sub-optimal topology problems, which limits their applications. In this paper, we propose a novel graph method CIFT to tackle these two problems by both model design and learning algorithm, achieving satisfying generalization on VI-ReID.

\noindent \textbf{Causal Inference in Computer Vision.}
The causal inference has recently aroused widespread interest, especially in the combination with computer vision~\cite{chalupka2014visual,lopez2017discovering,qi2020two,wang2020visual} to endow models with the ability to pursue the causal effect. 
Some works~\cite{chen2020counterfactual,niu2021counterfactual,rao2021counterfactual,tang2020long,tang2020unbiased,zhang2020counterfactual} utilized counterfactual to solve problems in various fields of computer vision.
Tang \etal.~\cite{tang2020long,tang2020unbiased} used counterfactual inference in scene graph generation and long-tailed classification to remove bias from training data with long-tailed distributions. Rao \etal.~\cite{rao2021counterfactual} used counterfactual training in fine-grained image recognition to tackle the bias of the spatial attention caused by the dataset. Niu \etal.~\cite{niu2021counterfactual} reduce the language bias in visual question answering by subtracting the direct language effect from the total causal effect.

Different from them, we focus on highlighting the affinity of the feature transfer module to address the sub-optimal topology structure due to the graph-based Re-ID model itself, rather than reducing the impact caused by biased data. 

%% file: contents/review.tex
\section{Delving into Graph-based Visible-Infrared  ReID}
\label{sec:3}

In this section, we investigate the influence of graph-based modules in the VI-ReID task. Specifically, we first give a brief review of graph-based VI-ReID methods. Then, we investigate why they suffered by bad generalization. Here, we take cm-SSFT~\cite{lu2020cross} and DDAG~\cite{ye2020dynamic} as examples for analysis. 

\subsection{Review of Graph-based VI-ReID Models}

The definition of VI-ReID is essentially a cross-modality retrieval task. So its formula can be written as follow: $\mathcal{R}=\mathcal{M}(q,G)$, where $\mathcal{M}$ is the Re-ID model, used to feedback the ranking list $\mathcal{R}$ between the given query sample $q$ and the gallery set $G$ whose modality is different with the query one. To achieve this pipeline, cm-SSFT~\cite{lu2020cross} and DDAG~\cite{ye2020dynamic} can be summarized as following:

\noindent
{\bf Step 1: Modality-invariant feature extraction.} Give an image $x_m$ whatever its modality $m$ ($m\in\{rgb,ir\}$), utilizing CNNs or other backbones to extract features $x$ for each sample. 

\noindent
{\bf Step 2: Feature enhancement.} Build affinities $A$ between all samples in $\{q,G\}$ on their given features $X$, where $A_{i,j}$ means the relationship between the $i$-th and the $j$-th images. After that, messages can be passed and transferred across different samples, leading to stronger features. It can be written as
\begin{equation}
F=A\cdot v(X), 
\end{equation}
where $v$ is a linear learnable function and $F$ stores the output features. This process is essentially equal to constructing a graph whose nodes are person features and edges are affinities and then propagating information based on that graph. 

\noindent
{\bf Step 3: Computing results.} After getting enhanced features, different kinds of outputs, e. g. person identities or ranking lists, can be derived.

\noindent
{\bf Step 4: Feature learning.} In the training stage, feature learning algorithms are added on both the backbone features $X$ and graph features $F$. The classification output $Y$ is derived by $F$ through a classification layer and a cross-entropy loss is used to train it, which makes features carrying identity information. 

The most priority of these graph-based modules is passing messages across samples, which mines the potential relationships between different person images. 
So, they can benefit both training~\cite{lu2020cross,ye2020dynamic} and inference~\cite{ye2020dynamic}. 
But we find that they are all suffering from bad generalization in VI-ReID.

\subsection{Analysis of Bad Generalization of Graph-based VI-ReID}
We summarize that the bad generalization of graph-based VI-ReID is caused by two following problems:

\noindent
{\bf Train-Test modality balance gap.} The train-test modality balance gap is caused by the difference of modality information ratio in the training and test stage. Specifically, in the training stage, both cm-SSFT~\cite{lu2020cross} and DDAG~\cite{ye2020dynamic} pass messages and transfer features on the batch data which includes an equal number of visible and infrared images. So the ratio of two modality information provided in training is $1:1$. But in inference, the available data is $\{q,G\}$ consisting of one query sample $q$ and a gallery set $G$. Here, the modality information ratio between two modalities is $1:N_G$, where $N_G$ is the size of the gallery set. It is clear that the modality information ratio of training and testing is quite different. This is the property of VI-ReID because VI-ReID utilizes the single query evaluation setting which means there is only one query sample available in the inference scenario. It is hard for the model trained on the balanced training data to generalize on the unbalanced inference scenario. The cm-SSFT~\cite{lu2020cross} also provides a series of experiments that demonstrate the unbalanced inference scenario can actually harm the generalization, which brings about 13.9\% Rank-1 and 9.1\% mAP drops corresponding to a balanced inference one.

\noindent
{\bf Sub-optimal topology structure.} The affinities $A$ computed by cm-SSFT~\cite{lu2020cross} and DDAG~\cite{ye2020dynamic} can indicate the relationships between different samples. So the $A$ can be interpreted as a kind of graph topology structure on the given data. But we argue that the structure learned by the existing graph VI-ReID modules are all sub-optimal because of the end-to-end joint learning.

Both cm-SSFT~\cite{lu2020cross} and DDAG~\cite{ye2020dynamic} add supervisions on backbone features and graph features simultaneously without any constraint on the affinities $A$, which hurts the generalization. It is common for the graph modules, like transformer~\cite{vaswani2017attention} or Graph Attention~\cite{velivckovic2017graph}, to train the $A$ in an end-to-end joint manner. But the situation is different here, the supervision on the backbone makes the backbone features $X$ discriminative in the training set. At this time, an $A$ with standard quality can make the final output belong to the feature learning constraints$^1$, so the structure $A$ cannot get much useful guidance. Without further supervision of $A$, it is hard for the graph module to capture the complex relationships between different samples.  

\renewcommand{\thefootnote}{1}
\footnotetext{Further proves could be seen in the supplementary.}

The above analyses reveal that the key to increasing the generalization of graph-based VI-ReID is reducing the modality balance gap between train-test and introducing additional constraints on $A$ in the end-to-end joint learning. Along this direction, we proposed a Counterfactual Intervention Feature Transfer module and show how this model is used to tackle these problems. 

%% file: contents/method.tex
\section{Counterfactual Intervention Feature Transfer}

\begin{figure*}[t]
  \centering
    \includegraphics[width=1.0\linewidth]{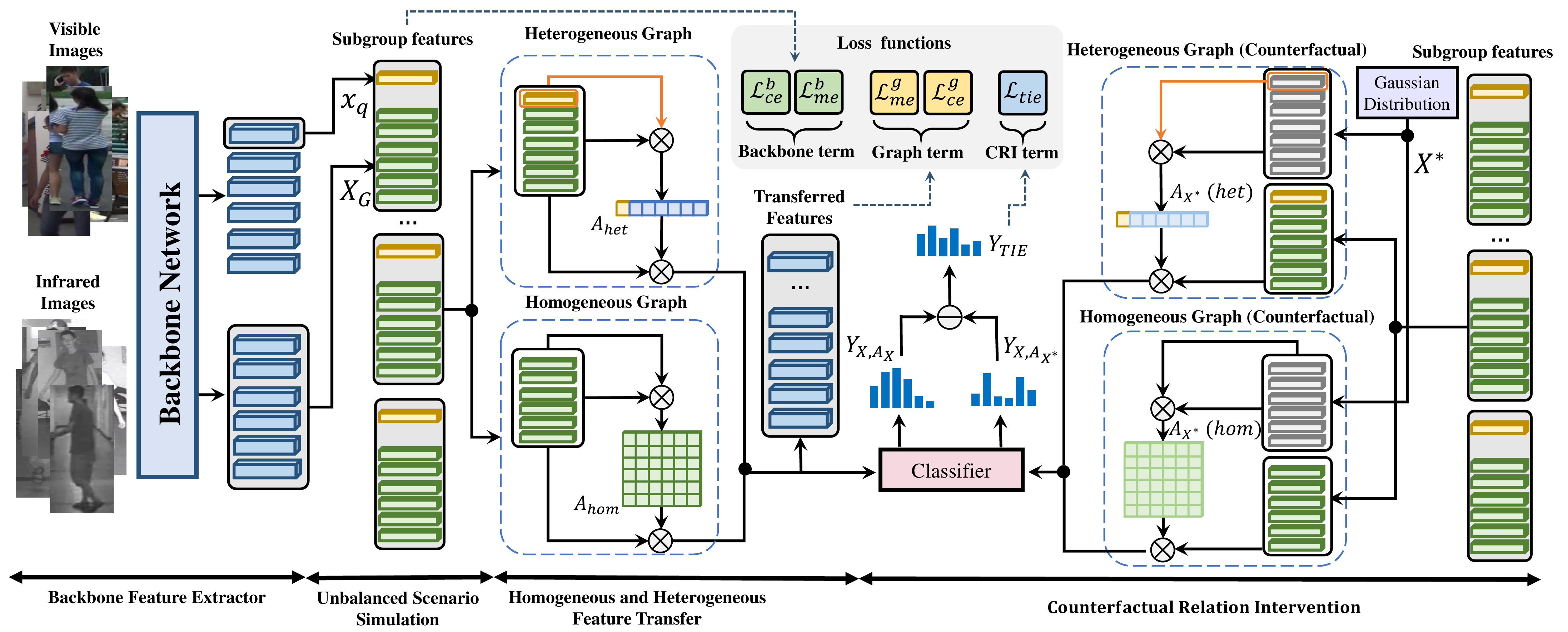}
    \caption{
    The Framework of the proposed Counterfactual Intervention Feature Transfer (CIFT). The Homogeneous and Heterogeneous Feature Transfer (H$^2$FT) module receives features from the backbone feature extractor and simulates unbalanced modality distributed scenarios. It builds homogeneous and heterogeneous message passing for better relationship feature learning under the unbalanced modality distribution. Then, the Counterfactual Relation Intervention (CRI) module calculates total indirect effects to highlight the role of the graph topology and leads to stronger results.}
    \label{fig:2}
  \hfill
\end{figure*}

The overview of our proposed Counterfactual Intervention Feature Transfer (CIFT) is shown in Fig.~\ref{fig:2}.
Visible images and infrared images are first fed into a pedestrian feature extractor to extract the instance-level features. Then these features are sent to the graph module Homogeneous and Heterogeneous Feature Transfer (H$^2$FT) to extract transferred features (§~\ref{sec:approach1}). Meanwhile, the Counterfactual Relation Intervention (CRI) is introduced to improve graph structure learning (§~\ref{sec:approach2}). The optimization and loss function details are shown in (§~\ref{sec:approach3}).

\subsection{Pedestrian representation backbone}

As shown in Fig.~\ref{fig:2}, our pedestrian representation backbone network is a weight sharing network for both modality data, which embeds the person images from different modalities to a same feature space. To make sure the representation ability of those features, a loss function $\mathcal{L}^b_{ce}$ including a classification loss and a cross-modality metric learning loss $\mathcal{L}^b_{me}$ is utilized to optimize the networks. Based on that, each person's image can have an initial representation.

\subsection{Homogeneous and Heterogeneous Feature Transfer}

For a given batch data including $N$ Visible and $N$ infrared images, we aim to simulate the unbalanced modality distributed scenario to train our model. The batch data is split into a series of groups and each group consists of a single image from one modality (seen as the query $q$) and $N$ other modality images (seen as the gallery set $G$). Specifically, each group can have one visible image with $N$ infrared images or one infrared image with $N$ visible images. Every group simulates the scenario with $1:N$ modality ratio which is similar in the inference setting (one query vs more galleries). So, our H$^2$FT trained under that can adapt on the single query inference without much influence on generalization. 

But, under that setting, the modality information is quite imbalanced. The sample seen as the query can only interact with $N$ images from other modality. And for the data seen as gallery, inter-modality interaction is trivial because it just introduces the information provided by the fixed query one, which is redundant even noisy. To avoid this problem, we provide a heterogeneous and a homogeneous graph module for the query data and gallery data separately like Fig.~\ref{fig:2} shows. Its equation can be written as: 

\begin{equation}
\begin{aligned}
    f_{q} =A_{het}\cdot [v(x_q),v(X_G)],\
    F_{G} =A_{hom}\cdot v(X_G),
\end{aligned}
\label{eq:t1}
\end{equation}
where $x_q$ and $X_G$ mean the query feature vector and the gallery features matrix respectively. $[\bullet,\bullet]$ means concatenation in the column dimension. The function $v(\cdot)$, a BNNeck~\cite{luo2019bag} with learnable weights, is used to enhanced the features. $f_q$ and $F_G$ mean the transferred features of the query and the gallery set.
$A$ means affinity matrix indicating the relationships between its corresponding samples. To achieve $A$, we first compute the similarity matrix based on the input features: 

\begin{equation}
\begin{aligned}
    \mathcal{S}_{ty}^{i,j} &= \exp{\frac{cos(v(x_i),v(x_j))}{\tau_{ty}}},\quad ty \in \{hom,het\}.
\end{aligned}
\label{eq:t}
\end{equation}
$\cos(\cdot,\cdot)$ is the cosine similarity function used to measure the similarity between samples. $\tau$ is the temperature parameter to adjust the smoothness of the total similarity distribution. We use different $\tau$ for the heterogeneous and homogeneous process because the intra-modality and inter-modality similarities are quite different.
To filter out the noisy relationships, we use near neighbor chosen function $\mathcal{T}(\bullet,k)$~\cite{lu2020cross} to keep the top-$k$ values in each row of similarity matrix: $\mathbf {S}' = \mathcal{T}(\mathbf {S},k)$. Finally, our affinity matrices are computed as follows:
\begin{equation}
\begin{aligned}
    A_{hom} = \mathbf {D}^{-1}_{hom}\cdot \mathbf {S}'_{hom},\,\,   A_{het} = \mathbf {D}^{-1}_{het}\cdot \mathbf {S}'_{het}.
\end{aligned}
\end{equation}
\rev{$\mathbf {D}$ is the diagonal degree matrix of $\mathbf {S}'$, with $D_{ii}=\sum_j S'_{ij}$. Thus, $\mathbf {D}^{-1}\mathbf {S}'$ row-normalizes the affinities.}

Finally, after achieving the final transferred features $F$, a classification layer derives the logits $Y$ and the cross-entropy loss $\mathcal{L}^g_{ce}$ is utilized to train $Y$ which is equal to maximizing likelihood to keep the features carrying richer identity information.
Then metric learning term $\mathcal{L}^g_{me}$ is added to the output to make features carry more discriminative information. 

Along that pipeline, the unbalanced scenario simulation makes the model adapt to the single query inference scenario. And the two kinds of message passing processes, heterogeneous and homogeneous, guide the query sample to interact with potential galleries while the gallery data only propagate messages across themselves, which preserves the nontrivial information interaction process in each group data, which is more suitable for tackling this unbalanced situation. 

\label{sec:approach1}
\subsection{Counterfactual Relation Intervention}

To add additional supervision of the affinities $A$ and keep the whole end-to-end training pipeline, we present to highlight the role of the graph topology structure in the total learning process. For this goal, we bring the tools of causal inference here. We first represent our H$^2$FT into a Structural Causal Model (SCM)~\cite{pearl2016causal,pearl2018book}, like Fig.~\ref{fig:1} (b) shows. $X\rightarrow A$ means the affinity computation and $X\rightarrow Y\leftarrow A$ is the message passing process (including the output computation).

It is obvious the process that deriving the output $Y$ from input $X$ can be seen as two types of effects: One is the direct effect $X\rightarrow Y$ and the other is an indirect one $X\rightarrow A\rightarrow Y$. The classification loss for our H$^2$FT, equal to maximizing the likelihood, would affect the two effects in an end-to-end manner so the $A$ in the indirect effect path cannot be enhanced sufficiently.  

To highlight the $A$ in the whole training process, we utilize the Total Indirect Effect (TIE) here. We first give its equation:
\begin{equation}
\begin{aligned}
    Y_{TIE} = Y_{X,A_X}- \mathbb{E}_{X^*}[Y_{X,A_{X^*}}].
\end{aligned}
\label{eq:ytie}
\end{equation}
$Y_{X,A_X}$ is the original output of our graph module, which means feeding forward the different sample features $X$ and computing their outputs. Note that the affinity matrix here is denoted as $A_X$ which means that affinities are computed based on the input features $X$.
$Y_{X,A_{X^*}}$ means computing the results by replacing original affinity $A_X$ to a intervened one $A_{X^*}$, where $X^*$ is the intervened inputs given manually. It is obvious that $Y_{X,A_{X^*}}$ cannot occur in the real world because features $X$ and affinities $A_{X^*}$ come from different inputs $X$ and $X^*$, which is called counterfactual intervention. So modification from $Y_{X,A_{X}}$ to $Y_{X,A_{X^*}}$ is equal to keep all potential variables fixed but only change the affinity $A$, which can show the pure effect introduced by $A$. We compute the expectation of that effect to get the more stable one. Intervened input features $X^*$ utilized to compute $A_{X^*}$ are sampled by a Gaussian distribution:
\begin{equation}
X^* = {X}_{\sigma}\cdot Z+{X}_{\mu},
\end{equation}
where $Z$ is the standard random vector whose dimension is same with features $X$. mean ${X}_{\mu}$ and stand deviation ${X}_{\sigma}$ are learned by the re-parameterization trick~\cite{kingma2013auto} in an end-to-end way. 

A cross-entropy loss is added to the TIE: $\mathcal{L}_{tie}=\mathcal{L}_{ce}(Y_{TIE})$.
Minimizing that cross-entropy loss is equal to maximizing the $Y_{TIE}$ on the prediction of the correct class, which guides the model to increase the gap between the original output and the counterfactual one. It is clear that the counterfactual classification results should be worse than the original one because the intervened affinities $A_{X^*}$ commonly do not match with inputs $X$. So an intuitive understanding about maximizing TIE is constraining the model to increase the difference between the outputs derived from good $A_X$ and bad $A_{X^*}$. Since other variables have been fixed, the model has to change the $A_X$ to increase the original results $Y_{X,A_{X}}$ for enhancing the gaps, leading to better training of affinity. 

\label{sec:approach2}

\subsection{Optimization}

The whole model is trained end-to-end and the total loss $\mathcal{L}_{total}$ of our method is defined as:$^2$
\begin{equation}
\begin{aligned}
\mathcal{L}_{total}=\underbrace{\mathcal{L}^b_{ce}+\mathcal{L}^b_{me}}_{\text{backbone term}}+\underbrace{\mathcal{L}^g_{ce}+\mathcal{L}^g_{me}}_{\text{graph term}}+\underbrace{\mathcal{L}_{tie}}_{\text{{CRI term}}}.
\end{aligned}
\end{equation}
\label{sec:approach3}
\label{sec:approach}
\renewcommand{\thefootnote}{2}
\footnotetext{The details about cross-modality metric learning loss \rev{$\mathcal{L}^b_{me}$ and $\mathcal{L}^g_{me}$} can be found in the supplementary.}

%% file: contents/experiments.tex
\section{Experiments}
\label{sec:exper}
\subsection{Datasets and Evaluation Protocol}

In this section, we conduct comprehensive experiments to evaluate our method on two public datasets, SYSU-MM01~\cite{wu2017rgb} and RegDB~\cite{nguyen2017person}.

\noindent{\bf SYSU-MM01} is the first large-scale benchmark dataset for Visible-Infrared ReID. It is collected by four visible and two infrared cameras, in both indoor and outdoor environments. The training set contains 395 identities with 22,258 visible and 11,909 infrared images while the test set contains 96 identities. Concretely, the query set contains 3,803 infrared images and the gallery set contains 301/3010 (single-shot/multi-shot) randomly selected visible images.

\noindent{\bf RegDB} is collected by a dual-camera system(a pair of aligned visible and thermal cameras). It contains 412 people, and each person has 10 visible and 10 far-infrared images.
The dataset is divided into training and test splits randomly, the images of 206 identities for training and the rest 206 identities for testing.

\noindent{\bf Evaluation Protocol.}
All the experiments follow the standard evaluation protocol in existing Visible-Infrared cross-modality ReID benchmarks. 
For SYSU-MM01, the original evaluation protocol~\cite{wu2017rgb} provides all-search and indoor-search modes for testing. 
Both search modes have two retrieval settings, single-shot and multi-shot.
For RegDB, we follow the widely used evaluation protocol in~\cite{nguyen2017person} which contains two modes for testing, Visible to Infrared test mode and Infrared to Visible test mode.
We evaluate our model on the 10 trials with different training/test splits to achieve stable performance.
For both datasets, the cumulative matching characteristics (CMC) and mean average precision (mAP) are adopted as evaluation metrics.

\subsection{Implementation Details}

We implement our approach with PyTorch~\cite{paszke2017automatic} on one NVIDIA Titan Xp GPU. Following the previous ReID methods~\cite{luo2019bag,ye2021deep}, we use ResNet-50~\cite{he2016deep} pre-trained on ImageNet as our backbone network. We change the stride of the last convolutional layer in the backbone to 1 and employ the Batch Normalization Neck~\cite{luo2019bag} as the embedding layer. Each person image is resized to commonly used 288 × 144 resolution. We also adopt the random cropping, random horizontal flipping and random erasing~\cite{zhong2020random} for data augmentation. The $k$ in the near neighbor chosen function is set to 4. \rev{$\tau_{hom}$ and $\tau_{het}$ in Eq.~\ref{eq:t} are set to 0.4 and 0.2 for the homogeneous and heterogeneous process, respectively.}
The whole model is trained for 120 epochs with the SGD optimizer. The learning rate gradually rises up by the warm-up scheme and decays by a factor of 10 at the 60th and 100th epochs. 
The batch size is set to 64, containing 32 visible and 32 infrared images from 8 identities. And each identity consists of 4 visible and 4 infrared images.

\begin{table*}[!t]
    \fontsize{7pt}{1.1em}\selectfont
    \centering
    \caption{Comparison of rank-1 accuracy (\%) and mAP accuracy(\%) with the state-of-the-art methods on SYSU-MM01 and RegDB. 
    (CIFT$^\dagger$ means we use backbone features for inference rather than the transferred graph features.)}
    \label{tab:comp}
    \begin{tabular}{r|c c|c c|c c|c c|c c|c c}
        \hline
        \multirow{4}{*}{Method}&
        \multicolumn{8}{c|}{SYSU-MM01~\cite{wu2017rgb}}&
        \multicolumn{4}{c}{RegDB~\cite{nguyen2017person}}
        \cr\cline{2-13}&
        \multicolumn{4}{c|}{All-search}&
        \multicolumn{4}{c|}{Indoor-search}&
        \multicolumn{2}{c|}{Visible to}&
        \multicolumn{2}{c}{Infrared to}
        \cr\cline{2-9}&
        \multicolumn{2}{c|}{Single-shot}&\multicolumn{2}{c|}{Multi-shot}&
        \multicolumn{2}{c|}{Single-shot}&\multicolumn{2}{c|}{Multi-shot}&
        \multicolumn{2}{c|}{Infrared}&
        \multicolumn{2}{c}{Visible}
        \cr\cline{2-13}&
        rank-1&mAP&rank-1&mAP&rank-1&mAP&rank-1&mAP&rank-1&mAP&rank-1&mAP
        \cr\hline
        Zero-Pad~\cite{wu2017rgb}
        &14.80&15.95&19.13&10.89&20.58&26.92&24.43&18.64&-&-&-&-\cr
        cmGAN~\cite{dai2018cross}
        &26.97&27.80&31.49&22.27&31.63&42.19&37.00&32.76&-&-&-&-\cr
        D$^2$RL~\cite{wang2019learning}
        &28.9&29.2&-&-&-&-&-&-&43.4&44.1&-&-\cr
        JSIA-ReID~\cite{wang2020cross}
        &38.1&36.9&45.1&29.5&43.8&52.9&52.7&42.7&48.5&49.3&48.1&48.9\cr
        AlignGAN~\cite{wang2019rgb}
        &42.4&40.7&51.5&33.9&45.9&54.3&57.1&45.3&57.9&53.6&56.3&53.4\cr
        AGW~\cite{ye2021deep}
        &47.5&47.65&-&-&54.17&62.97&-&-&70.05&66.37&-&-\cr
        cm-SSFT~\cite{lu2020cross}
        &61.6&63.2&63.4&62.0&70.5&72.6&73.0&72.4&72.3&72.0&71.0&71.7\cr
        cm-SSFT(sq)
        &47.7&54.1&-&-&57.4&59.1&-&-&65.4&65.6&63.8&64.2\cr
        DDAG~\cite{ye2020dynamic}
        &54.75&53.02&-&-&61.02&67.98&-&-&69.34&63.46&68.06&61.80\cr
        HC~\cite{zhu2020hetero}
        &56.96&54.95&62.09&48.02&59.74&64.91&69.76&57.81&-&-&-&-\cr
        CIMA~\cite{zhao2021joint}
        &57.2&59.3&60.7&52.6&66.6&74.7&73.8&68.3&78.8&69.4&77.9&69.4\cr
        HCT~\cite{liu2020parameter}
        &61.68&57.51&-&-&63.41&68.17&-&-&91.05&83.28&89.30&81.46\cr
        MCLNet~\cite{hao2021cross}
        &65.4&61.98&-&-&72.56&76.58&-&-&80.31&73.07&75.93&69.49\cr
        SMCL~\cite{wei2021syncretic}
        &67.39&61.78&72.15&54.93&68.84&75.56&79.57&66.57&83.93&79.83&83.05&78.57\cr
        MPANet~\cite{wu2021discover}
        &70.58&68.24&75.58&62.91&76.74&80.95&84.22&75.11&83.7&80.9&82.8&80.7\cr
        \hline
        CIFT$^\dagger$(Ours)
        &71.77&67.64
        &78.00&62.46
        &78.65&82.11
        &86.97&77.03
        &{\bf92.17}&86.96
        &90.12&84.81\cr
        CIFT(Ours)&{\bf74.08}&{\bf74.79}&{\bf79.74}&{\bf75.56}&{\bf81.82}&{\bf85.61}&{\bf88.32}&{\bf86.42}&91.96&{\bf92.00}&{\bf90.30}&{\bf90.78}\cr
        \hline
    \end{tabular}
\end{table*}

\subsection{Comparison with State-of-the-art Methods}
\label{subsec:sotas}

In this part, we compare our proposed method CIFT with state-of-the-art (SOTA)
visible-infrared person Re-ID approaches, including Zero-Pad~\cite{wu2017rgb}, cmGAN~\cite{dai2018cross}, D$^2$RL~\cite{wang2019learning}, JSIA-ReID~\cite{wang2020cross}, AlignGAN~\cite{wang2019rgb},
AGW~\cite{ye2021deep}, cm-SSFT~\cite{lu2020cross}, 
DDAG~\cite{ye2020dynamic}, HC~\cite{zhu2020hetero}, 
CIMA~\cite{zhao2021joint}, HCT~\cite{liu2020parameter}, MCLNet~\cite{hao2021cross}, SMCL~\cite{wei2021syncretic} and MPANet~\cite{wu2021discover}.

\noindent {\bf Comparison and Analysis. }
The experimental results are shown in Table~\ref{tab:comp} and the proposed method outperforms the existing SOTAs on both datasets. In SYSU-MM01 dataset, our CIFT achieves 74.08\% rank-1 accuracy and 74.79\% mAP accuracy, which surpasses  MPANet~\cite{wu2021discover} by 3.50\% on rank-1 accuracy and 6.55\% on mAP accuracy in the most challenging single-shot all search mode. Even our CIFT$^\dagger$ (only uses graph module in training and utilizes the backbone features in inference) outperforms MPANet by 1.19\% on rank-1 accuracy. In another popular public RegDB dataset, whether in 'infrared to visible' mode or 'visible to infrared' mode, our CIFT$^\dagger$ still achieves the highest scores. The average performances in the two modes are 91.15\% rank-1 accuracy and 85.89\% mAP accuracy, which surpasses MPANet by a large gain of about 7.90\% on rank-1 accuracy and 5.09\% on mAP accuracy. This is because the learning of transferred features introduces additional supervision to the model, so that the features of the backbone network are also enhanced.

For the multi-shot setting, the mAP accuracy of all other methods will drop significantly compared with the single-shot evaluation because the model is required to retrieve more positive targets in the multi-shot setting. So it is more challenging for the model to find out all potential targets. 
But our CIFT is not suffering that bad phenomenon even can achieve better results, which shows that our method is qualitatively different from other methods. When the scale gallery size is larger, our model can extract richer and more discriminate relation features so that the model is more robust to gallery size even benefited by the larger scale one. Compared with cm-SSFT~\cite{lu2020cross} which also obtain relationship from the gallery set, our CIFT improves the mAP accuracy by 0.77\% as the gallery size increases (multi-shot versus single-shot), while the cm-SSFT reduces the mAP accuracy by 1.2\%. This is because we utilize CRI to highlight the role of topology structure in the whole training process. The affinities are much more accurate so that the final representation is much stronger. 

\subsection{Comparison with Multi-gallery Matching Methods}
\label{subsec:gallery}

To demonstrate the superiority of our CIFT to other graph and post-process VI-ReID methods, we compare it with other multi-gallery matching methods, including cm-SSFT~\cite{lu2020cross}, k-reciprocal rerank~\cite{zhong2017re}, and GNN rerank~\cite{zhang2020understanding}. We evaluate these methods with our CIFT on different backbone networks including AGW~\cite{ye2021deep}, HC~\cite{zhu2020hetero}, HCT~\cite{liu2020parameter}, and our backbone network, to show the generality of our method under different level baselines. For a fair comparison,
we also search the best hyper-parameters for these multi-gallery matching methods, so that they can fit the backbones well. Please note that, to adapt these backbone features format, cm-SSFT is set as only using the shared feature transfer. 

Our results are shown in the 5th line in Table~\ref{tab:gallery} and 1st$\sim$3rd lines represent different widely used post-processing ways, combining different multi-gallery matching methods with the trained backbone features directly in inference. Comparing with the strongest post-process GNN rerank~\cite{zhang2020understanding}, we achieve averaged 5.12\% rank-1 and 4.03\% mAP gains on all given backbones. 
Further, we also compare the single query cm-SSFT (the 4th row) who does additional feature transfer learning corresponding to the aforementioned post-process methods. It only achieves comparable results with GNN rerank in the 3rd line and does not show much more priority of its graph learning. That is because cm-SSFT trains in the case of balanced modality but transferred to modal unbalanced inference scenario, which hurts its generalization. So, with the ability to tackle the inference under unbalanced modality distribution, it is common for CIFT to suppress the cm-SSFT by a large margin. 
%
%
The results in Table~\ref{tab:gallery} show that our method achieves the best performance on all backbone networks, bringing average improvements of 5.03\% on rank-1 accuracy and 9.50\% on mAP accuracy. This proves that our method is compatible with various backbone networks and can achieve effective improvement.

\begin{table}[!t]
    \fontsize{8pt}{1.3em}\selectfont
    \centering
    \caption{
    Compared with other methods that use gallery set information in the inference stage on same backbone networks, \ie AGW~\cite{ye2021deep}, HC~\cite{zhu2020hetero}, HCT~\cite{liu2020parameter} and our backbone. We report the rank-1 accuracy (\%) and the mAP accuracy (\%) on the SYSU-MM01 single-shot all search mode. 'train' and 'test' in 'strategy' means the training methods and test methods separately}
    \label{tab:gallery}
    \begin{tabular}{c|l|c c|c c|c c|c c}
        \hline
        \multirow{2}{*}{Row}&
        \multirow{2}{*}{
        \diagbox[width=8em]{Strategy}{backbone}}&
        \multicolumn{2}{c|}{AGW~\cite{ye2021deep}}&
        \multicolumn{2}{c|}{HC~\cite{zhu2020hetero}}&
        \multicolumn{2}{c|}{HCT~\cite{liu2020parameter}}&
        \multicolumn{2}{c}{Our backbone}
        \cr&&rank-1&mAP&rank-1&mAP&rank-1&mAP&rank-1&mAP\cr\hline
        1&backbone&47.22&47.78&54.52&54.06&61.05&56.97&70.49&66.58\cr
        2&k-reciprocal~\cite{zhong2017re}&47.63&51.81&53.91&60.03&62.33&62.05&71.47&72.49\cr
        3&GNN rerank~\cite{zhang2020understanding}&46.96&52.01&55.05&59.79&60.52&62.72&70.40&73.21\cr
        4&cm-SSFT~\cite{lu2020cross}&48.65&51.76&56.17&61.36&62.27&61.85&69.92&71.77\cr\hline
        5&{\bf CIFT}&{\bf52.12}&{\bf56.92}&{\bf61.03}&{\bf64.05}&{\bf66.18}&{\bf68.09}&\bf{74.08}&\bf{74.79}\cr\hline
    \end{tabular}
\end{table}

\begin{table}[!t]
    \small
    \centering
    \caption{Ablation study on SYSU-MM01. The important modules of the proposed CIFT, \ie H$^2$FT and CRI are analyzed under different settings.}
    \label{tab:abl}
    \begin{tabular}{c|c c c c|c c}
        \hline
        \multirow{2}{*}{Row}&\multirow{2}{*}{GFT}&\multirow{2}{*}{UBS}&\multirow{2}{*}{H$^2$G}&\multirow{2}{*}{CRI}
        &\multicolumn{2}{c}{SYSU-MM01}
        \cr\cline{6-7}
        &&&&&rank-1&mAP\cr\hline
        1&-&-&-&-&70.49&66.58\cr
        2&\checkmark&-&-&-&72.01&72.12\cr
        3&\checkmark&\checkmark&-&-&72.90&71.97\cr
        4&\checkmark&\checkmark&\checkmark&-&72.29&73.79\cr
        5&\checkmark&\checkmark&\checkmark&\checkmark&74.08&74.79\cr\hline
    \end{tabular}
\end{table}

\begin{table}[!t]
    \fontsize{9pt}{1.1em}\selectfont
    \centering
    \caption{Affinity quality statistics on the SYSU-MM01 test set. The value (\%) represents the {\bf average error ratio} of the affinity matrix in the entire test set.}
    \label{tab:err}
    \begin{tabular}{l|c|c|c|c}
        \hline
        \multirow{2}{*}{Method}&
        \multicolumn{2}{c|}{All-search}&
        \multicolumn{2}{c}{Indoor-search}
        \cr\cline{2-5}&Single-shot&Multi-shot&Single-shot&Multi-shot
        \cr\hline
        w/o CRI&5.16&3.95&6.54&6.15\cr
        w/ CRI&3.90&2.76&5.03&4.66\cr\hline
    \end{tabular}
\end{table}

\subsection{Ablation Study}
\label{subsec:ablation}

In this section, we conduct ablation studies to prove the effectiveness of each module of the proposed CIFT, \ie H$^2$FT and CRI. All ablation experiments are performed on our baseline backbone in the single-shot all search mode of the large-scale dataset SYSU-MM01. 
The results are shown in Table~\ref{tab:abl}.

\noindent{\bf Effectiveness of H$^2$FT.}
In the proposed CIFT, we introduce a graph-based feature transfer module H$^2$FT to tackle the train-test modality balance gap. 
To evaluate the effectiveness of each detailed part in H$^2$FT, 
We split the H$^2$FT into three parts: Graph Feature Transfer (GFT), UnBalanced Scenario simulation (UBS) and Homogeneous\&Heterogeneous Graph module (H$^2$G). GFT is a simple graph module baseline proposed by ourselves, which is used to show the gain introduced by the message passing in the graph module itself. Its details can be seen in the supplementary and we try ourselves to keep other variables not changing. Its performance in the 2nd row shows that the feature transfer can bring about 5.5\% gains in mAP. To train the model suitable for the unbalanced inference scenario, we add the UBS on it. But we find that the performance on mAP has a little drop. We think that is caused by the model design who is not fit the unbalanced data. Now, we add the H$^2$G back in the 3rd line, equal to the complete H$^2$FT, and achieve additional 1.67\% mAP gains. proving the effectiveness of our H$^2$FT. 

\noindent{\bf Effectiveness of CRI.}
The proposed CRI algorithm utilizes the counterfactual intervention to highlight the role of the topology to tackle the sub-optimal topology structure problem. As shown in Table~\ref{tab:abl}, CRI algorithm brings improvements of 1.79\% rank-1 accuracy and 1.00\% mAP accuracy without any computational costs in the inference stage.
In addition, we also do another quantitative analysis of the CRI to demonstrate its contribution. Specifically, we first compute the ideal affinity matrix by the ground-truth label, which uses 1 to indicate the positive pair and 0 as the negative one. And then for the affinity matrix computed by our model, we define the top-4 results in each row as positive. After that, we compute the averaged error rate between the predicted affinities with the ground truths in the entire test set. The results are shown in Table~\ref{tab:err}. In the single-shot all-search mode, the model without CRI gets for 5.16\% error ratio while the model with CRI achieves 3.90\%. In other test modes, introducing CRI also significantly reduces the error ratios. This reflects that our CRI has learned a better structure which is more close to the ground-truth one. 

%% file: contents/conclusion.tex
\section{Conclusion}
\label{sec:conclu}
We propose a Homogeneous and Heterogeneous Feature Transfer (H$^2$FT) module with a Counterfactual Relation Intervention (CRI) learning method to tackle the Visible-Infrared Person Re-identification. The H$^2$FT consists of two types of graph modules that can handle the train-test modality balance gap that the previous graph-based model suffered. And CRI introduces the causal inference tool to tackle the sub-optimal topology structure problem and makes our method more generalized. 

%% file: contents/prove.tex
\section{Further proof about sub-optimal topology structure}
The motivation of the Counterfactual Relation Intervention (CRI) is that joint learning of graph inputs $X$ and outputs $Y$ leads to bad affinities $A$. Here, we do some toy example experiments to demonstrate that phenomenon.

\begin{figure*}[t]
    \centering
    \includegraphics[width=0.9\linewidth]{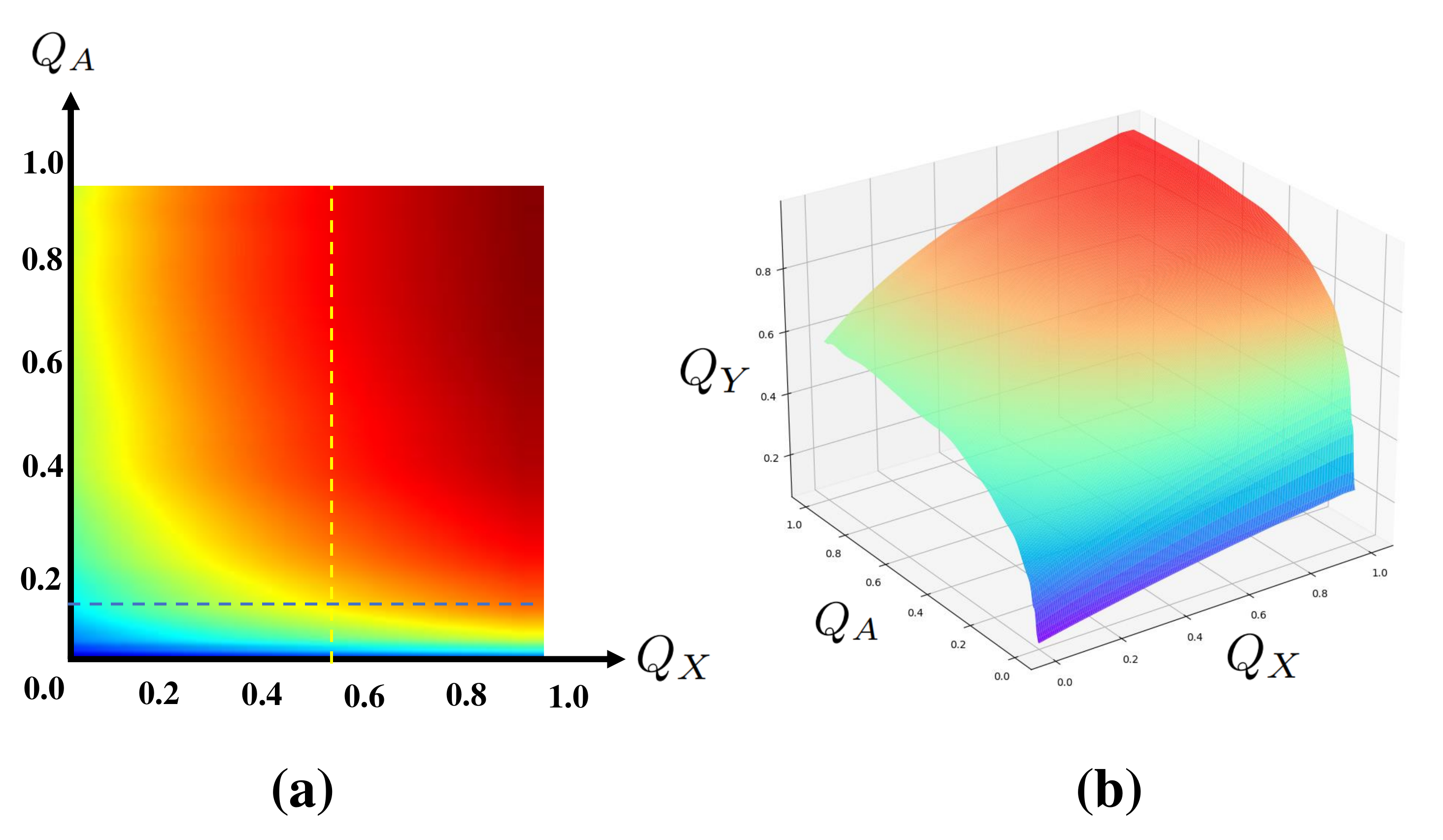}
    \vspace{-0.5cm}
    \caption{(a) shows the relationships between $Q_A$, $Q_X$ and $Q_Y$. \rev{The blue dash line fixes a low $Q_A$ and shows the changes of $Q_Y$ introduced by $Q_X$. The yellow dash line fixes the $Q_X$ value computed on our baseline on the SYSU-MM01 benchmark and shows the changes from $Q_A$ to $Q_Y$.}
    (b) is the 3D surface version of (a). }
    \label{fig:prove}
\end{figure*}

We aim to investigate the relationships of $A$ with the quality of $X$ and $Y$. For example, if the quality of $X$ and $Y$ are good (meaning both graph and backbone features trained well), how about the quality of $A$. We firstly define the metric to evaluate features and affinities. \rev{For the $i$-th sample, let $\mathcal{P}_i$ and $\mathcal{N}_i$ denote the positive and negative index sets, respectively. For a good learned representation, the distance to positive samples should be smaller than the distance to negative samples. Therefore, we define the average margin of input and output features as}
\begingroup
\color{black}
\begin{equation}
\begin{aligned}
M_i^x
&= \frac{1}{|\mathcal{P}_i||\mathcal{N}_i|}
\sum_{j \in \mathcal{P}_i}\sum_{k \in \mathcal{N}_i}
\left[d_{\cos}(x_i,x_k)-d_{\cos}(x_i,x_j)\right],\\
M_i^y
&= \frac{1}{|\mathcal{P}_i||\mathcal{N}_i|}
\sum_{j \in \mathcal{P}_i}\sum_{k \in \mathcal{N}_i}
\left[d_{\cos}(y_i,y_k)-d_{\cos}(y_i,y_j)\right],
\end{aligned}
\end{equation}
\endgroup
\rev{where $d_{\cos}(a,b)=1-\cos(a,b)$. These equations compute the averaged distance margin between positive pairs and their corresponding negative ones. We statistic the averaged margin as the quality metrics:}
\begingroup
\color{black}
\begin{equation}
    Q_X = \frac{1}{N} \sum_{i=1}^{N} M_i^x,\quad
    Q_Y = \frac{1}{N} \sum_{i=1}^{N} M_i^y,
\end{equation}
\endgroup
where $N$ is the number of features, equal for the graph one and the backbone one. And we also propose a metric to evaluate affinities $A$:
\begingroup
\color{black}
\begin{equation}
    Q_A =
    \frac{1}{N}\sum_{i=1}^{N}
    \frac{1}{|\mathcal{P}_i||\mathcal{N}_i|}
    \sum_{j \in \mathcal{P}_i}\sum_{k \in \mathcal{N}_i}
    \mathbf{1}[A_{i,j} > A_{i,k}].
\end{equation}
\endgroup
\rev{where $\mathbf{1}[\cdot]$ is the indicator function. That equation means that good affinities should include larger positive similarities than corresponding negative ones. So this $Q_A$ metric measures the ratio of positive-negative sample pairs belonging to that constraint.}

After that, we randomly generate a series of triplets $\{X,A,Y\}$ and control their quality carefully. And then, we evaluate the quality of $Q_Y$ and draw these data in Figure~\ref{fig:prove} and see how do the $Q_X$ and $Q_A$ affect $Q_Y$. To achieve stable results, we reproduce that statistic 100 times and compute mean results. From Figure~\ref{fig:prove} (a) (bird of view figure), it obvious that $A$ cannot learn sufficient when the inputs features are trained well. A low-quality affinities $A$ (low $Q_A$, e.g. about 0.1, the blue dash line in Figure~\ref{fig:prove} (a)) can also get a good transferred features $Y$ (high $Q_Y$) as long as having a high-quality input representation $X$ (high $Q_X$). It proves that if the $A$ is not good, the graph output features $Y$ can also have a good representation abilities because of good $X$. So the training of $X$ can relax the constraint of $A$.
And as we know, in the training set of ReID task, features also trained well. So the range of $A$ could be flexible. \rev{We compute the baseline feature quality on SYSU-MM01 and analyze the $Q_A$ influence to $Q_Y$ on this fixed $Q_X$ value} (yellow dash line in Figure~\ref{fig:prove} (a)). And we find that, the range of $A$ is extreme big. As long as the $Q_A$ upper than 0.16, the outputs of $Y$ can achieve satisfied quality, larger than 0.7.
It shows that well-trained $X$ and $Y$ can lead to a flexible range of $A$, bringing sub-optimal topology structure.

%% file: contents/loss.tex
\section{Feature learning loss function details}

For both backbone and graph module features, we add cross-entropy losses to train the features include identity information:
\begingroup
\color{black}
\begin{equation}
\begin{aligned}
    p_i^b &= \mathrm{Softmax}(W_bF_b^i),\quad
    p_i^g = \mathrm{Softmax}(W_gF_g^i),\\
    \mathcal{L}^b_{ce}
    &= \mathbb{E}_{i}[-\log p_i^b[y_i]],\quad
    \mathcal{L}^g_{ce}
    = \mathbb{E}_{i}[-\log p_i^g[y_i]],
\end{aligned}
\label{eq:ce_loss}
\end{equation}
\endgroup
\rev{where $y_i$ is the ground-truth identity label of the $i$-th sample, and $p_i^b[y_i]$ and $p_i^g[y_i]$ denote the predicted probabilities of the ground-truth category. $W_b$ and $W_g$ are learnable parameters of classification layers.}
Except that, we design a new metric learning loss called Heterogeneous Center Contrastive (HCC) loss:
\begingroup
\color{black}
\begin{equation}
\begin{aligned}
    \mathcal{L}_{hcc}(I,C,\mathcal{D})
    =\mathbb{E}_{i}\{\mathcal{D}(I^i,C_i^+)
    +\mathbb{E}_{k}[\text{max}(\rho_\mathcal{D}-\mathcal{D}(I^i,C_{i,k}^-),0)]\},
\end{aligned}
\label{eq:hcc}
\end{equation}
\endgroup
\rev{where $I$ is the input features and $C$ is their heterogeneous center. $C_i^+$ is computed by the positive samples of the $i$-th sample from the other modality in the current batch, and $C_{i,k}^-$ is computed by the negative samples of the $k$-th negative identity from the other modality. This metric learning loss essentially puts the features close to their corresponding heterogeneous positive centers and puts them away from the heterogeneous negative centers.}
We add this loss to both graph features and the backbone features. Their metric learning losses are:
\begingroup
\color{black}
\small
\begin{equation}
\begin{aligned}
    \mathcal{L}^b_{me}=\underbrace{\mathcal{L}_{hcc}(F^b,C^{F^b},Eu)}_{\text{feature-level}}+\underbrace{\mathcal{L}_{hcc}(p^b,C^{p^b},KL)}_{\text{probability-level}},\quad \mathcal{L}^g_{me}=\underbrace{\mathcal{L}_{hcc}(p^g,C^{p^g},KL)}_{\text{probability-level}},
\end{aligned}
\label{eq:me_loss}
\end{equation}
\endgroup
\rev{where the feature-level HCC loss aims to guide features embed in the Euclidean ($Eu$) space well. And the probability-level one puts KL-divergence ($KL$) constraints on the classification probability distribution.} We set $\rho_{Eu}=0.6$ and $\rho_{KL}=6$, respectively.

%% file: contents/gft.tex
\section{Graph Feature Transfer (GFT) details}
In the original paper, we split the H$^2$FT into three parts in the ablation study: Graph Feature Transfer (GFT), UnBalanced Scenario simulation (UBS) and Homogeneous$\&$Heterogeneous Graph module (H$^2$G). Here we further introduce the details of GFT. Similar to the original H$^2$FT, it can be defined as:

\begingroup
\color{black}
\begin{equation}
\begin{aligned}
    F =A\cdot v(X),
\end{aligned}
\end{equation}
\endgroup
where $X$, $F$ and $A$ mean input features matrix, transferred features and affinity matrix respectively. $X$ is the whole batch data consisting of $N$ rgb and $N$ infrared modality data (1 query with $N_G$ galleries in inference).
$A$ is computed by the following equations:
\begin{equation}
\begin{aligned}
    A = \mathbf {D}^{-1}\cdot \mathbf {S}',\quad \mathbf {S}' = \mathcal{T}(\mathbf {S},k),
\end{aligned}
\end{equation}
\rev{where $\mathbf {D}$ is a diagonal degree matrix with $D_{ii}=\sum_j S'_{ij}$, so $\mathbf {D}^{-1}\mathbf {S}'$ row-normalizes the affinities.}
The similarity matrix is computed as
\begin{equation}
\begin{aligned}
    \mathcal{S}^{i,j} &= \exp{\frac{\cos(v(x_i),v(x_j))}{\tau}}.
\end{aligned}
\end{equation}
$A$ is computed by the full $X$, which is different from our H$^2$FT. It is obvious that GFT transfers features in all batch in the training stage, suffering from the train-test modality balance gap.

%% file: contents/vis.tex
\section{More visualization results}
\begin{figure*}[h!]
    \centering
    \includegraphics[width=1.0\linewidth]{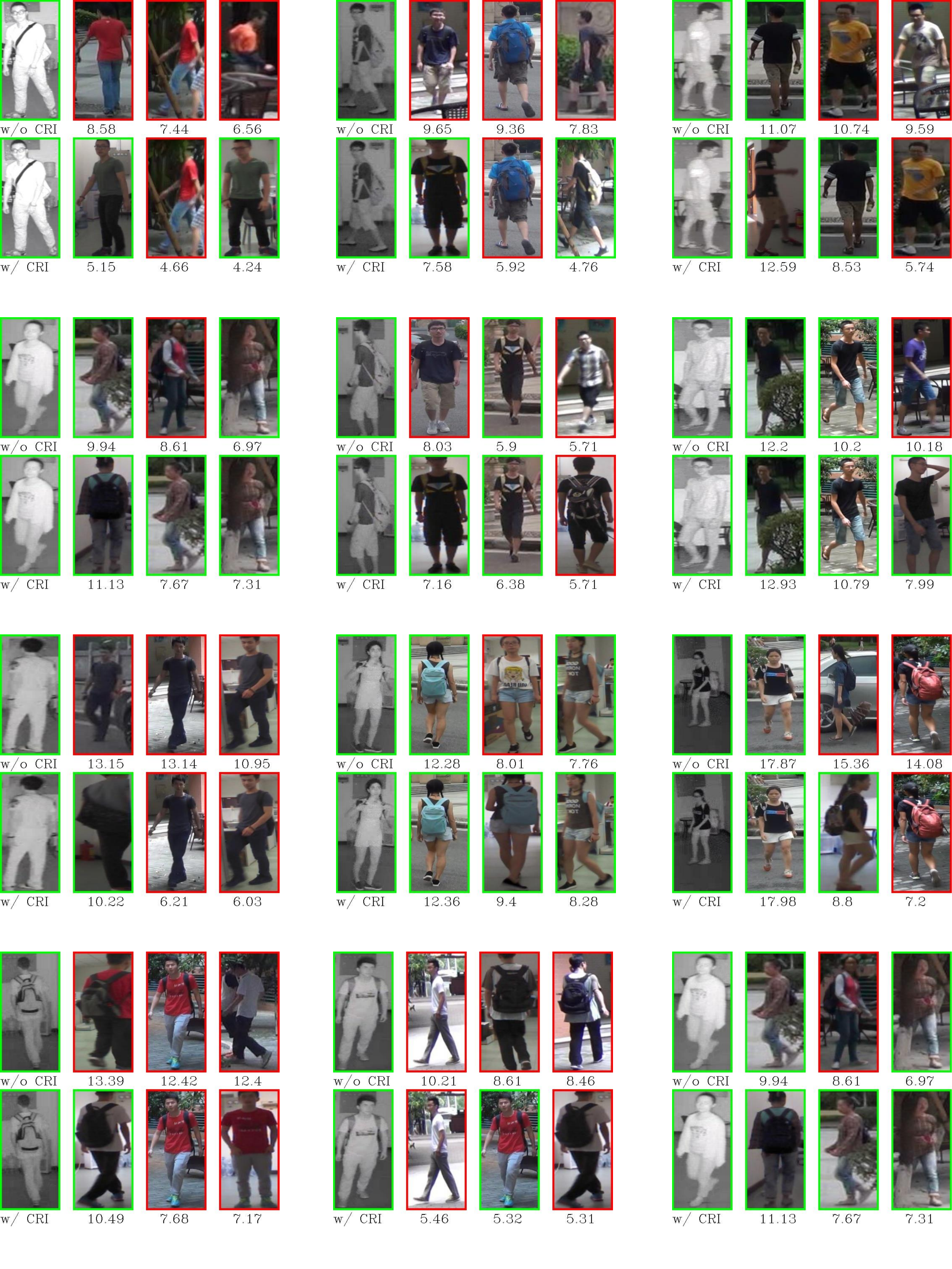}
    \caption{CRI leads to more positive samples for graph message passing.}
    \label{fig:s1}
\end{figure*}

\begin{figure*}[h!]
    \centering
    \includegraphics[width=1.0\linewidth]{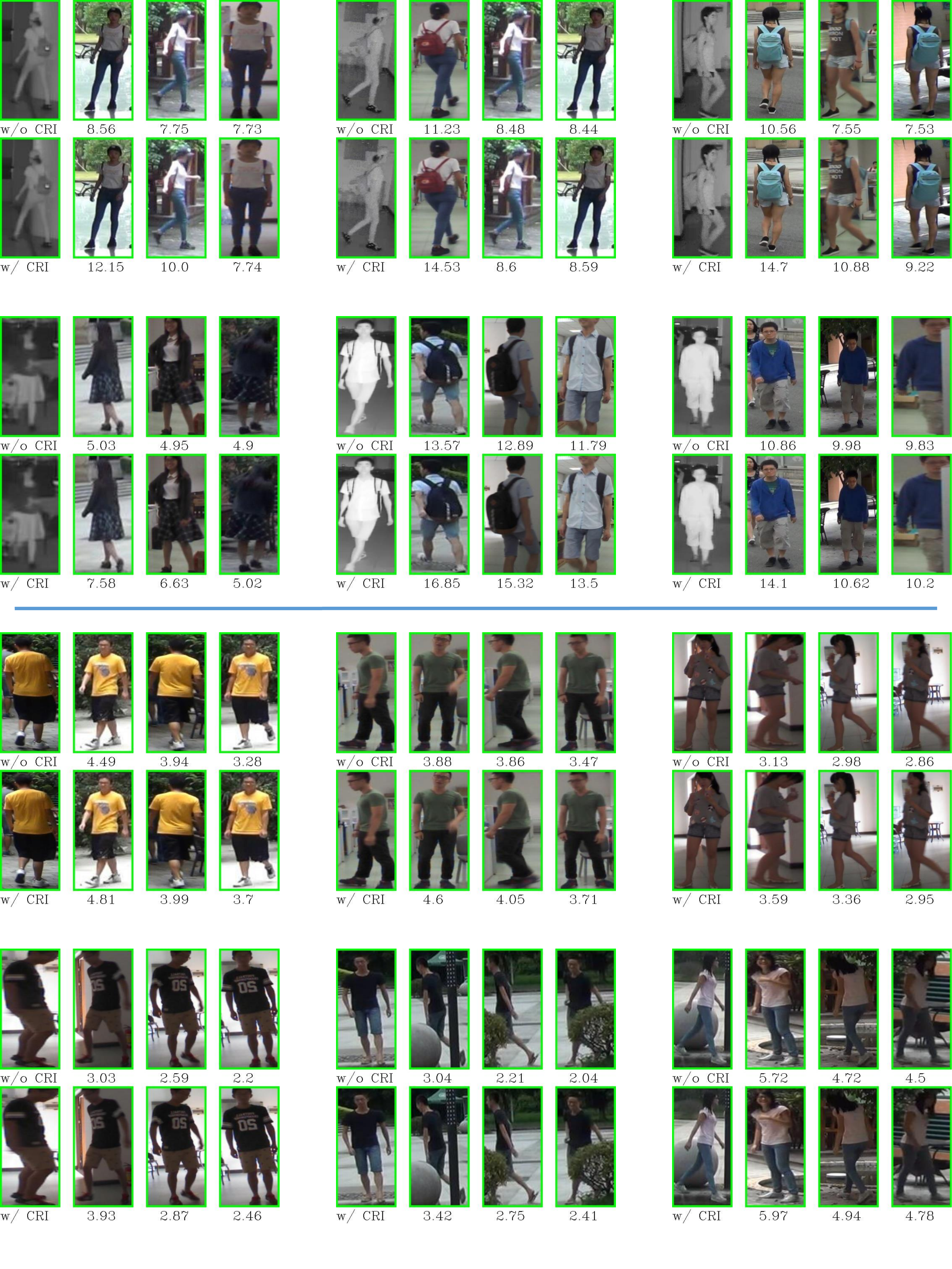}
    \caption{CRI can enlarge positive affinities between easy sample pairs.}    
    \label{fig:s2}
\end{figure*}

\begin{figure*}[h!]
    \centering
    \includegraphics[width=1.0\linewidth]{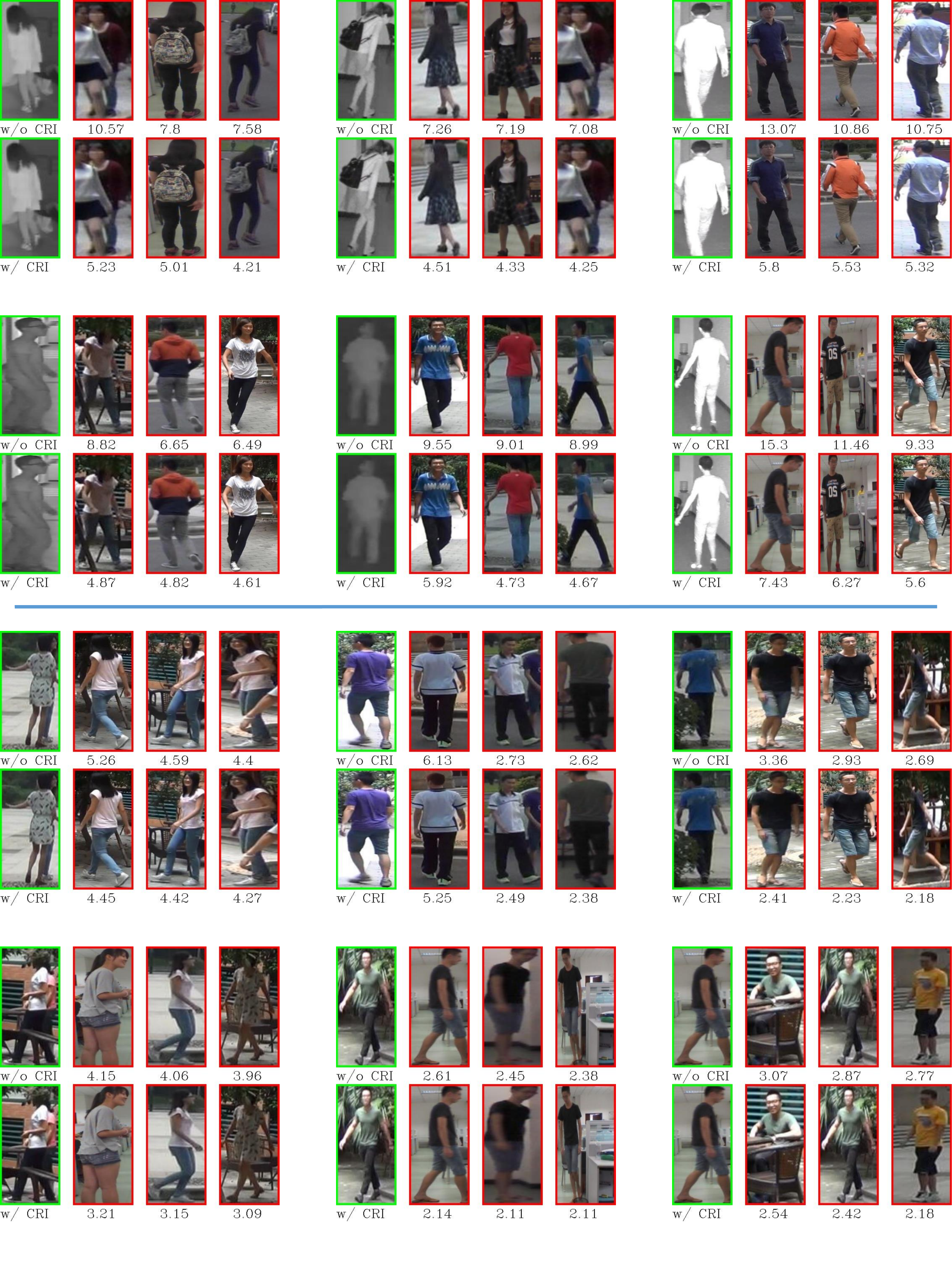}
    \caption{CRI can suppress negative affinities between hard samples.}  
    \label{fig:s3}
\end{figure*}

\begin{figure*}[h!]
    \centering
    \includegraphics[width=1.0\linewidth]{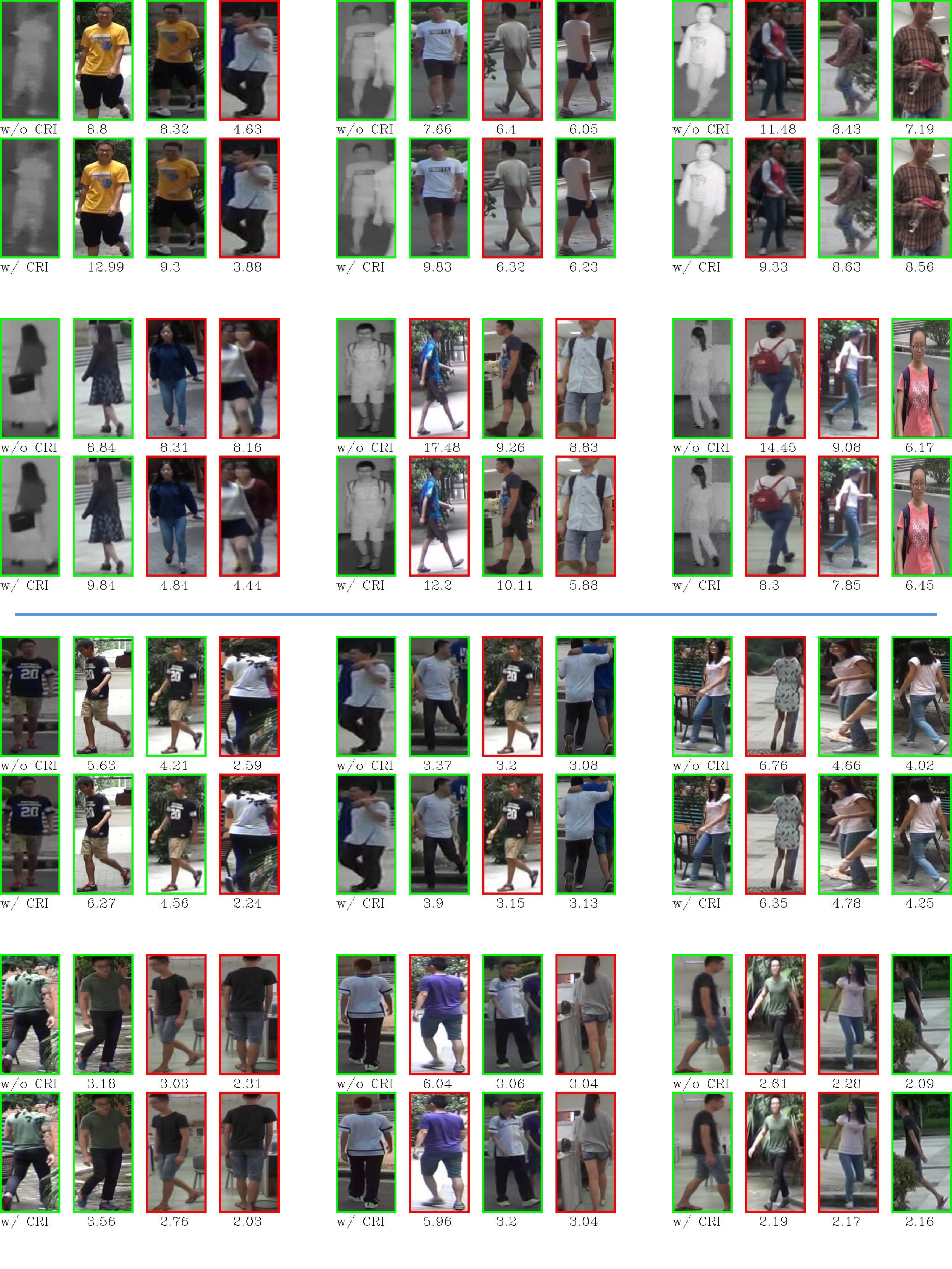}
    \caption{CRI can enlarge positive affinities and suppress negative ones simultaneously.}  
    \label{fig:s4}
\end{figure*}

We give more CRI visualizations about different views under the SYSU-MM01 single-shot all search mode in Fig.~\ref{fig:s1}, Fig.~\ref{fig:s2}, Fig.~\ref{fig:s3} and Fig.~\ref{fig:s4}. 

The visualizations all mean affinities with and without CRI. For each group, the first image is the sample preparing to interact and others are the top-3 similar samples of the first one. The green boxes represent correct matches, and the red boxes represent incorrect matches.
The results more intuitively show the effectiveness of our method for improving affinity.

